\documentclass[12pt]{article}
\usepackage{amsmath}
\usepackage{amssymb}
\usepackage{graphicx}
\usepackage{hyperref}
\usepackage{geometry}
\usepackage{subcaption}
\usepackage{placeins} 

\title{Partition of Unity Physics-Informed Neural Networks (POU-PINNs): \\ An Unsupervised Framework for Physics-Informed Domain Decomposition and Mixtures of Experts}
\author{
    Arturo Rodriguez$^1$, Ashesh Chattopadhyay$^2$, Piyush Kumar$^1$, \\Luis F. Rodriguez$^3$, Vinod Kumar$^4$\\
    \small $^1$The University of Texas at El Paso \\
    \small $^2$University of California, Santa Cruz \\
    \small $^3$Clarkson University \\
    \small $^4$Texas A\&M University in Kingsville
}
\date{}

\begin{document}

\maketitle

\begin{abstract}
Physics-informed neural networks (PINNs) commonly address ill-posed inverse problems by uncovering unknown physics. This study presents a novel unsupervised learning framework that identifies spatial subdomains with specific governing physics. It uses the partition of unity networks (POUs) to divide the space into subdomains, assigning unique nonlinear model parameters to each, which are integrated into the physics model. A vital feature of this method is a physics residual-based loss function that detects variations in physical properties without requiring labeled data. This approach enables the discovery of spatial decompositions and nonlinear parameters in partial differential equations (PDEs), optimizing the solution space by dividing it into subdomains and improving accuracy. Its effectiveness is demonstrated through applications in porous media thermal ablation and ice-sheet modeling, showcasing its potential for tackling real-world physics challenges.
\end{abstract}

\section{Introduction}
Physics-informed neural networks (PINNs) are a robust framework designed to solve partial differential equations (PDEs) through optimization methods for forward problems and data-driven discovery for inverse problems \cite{Raissi2019, Karniadakis2021}. While effective, PINNs face challenges in addressing parametrized PDEs with highly variable subdomains or discontinuities \cite{Shukla2021}. Standard PINNs struggle with convergence speed and accuracy because they cannot efficiently represent domain-specific features or variations \cite{Cai2021, Wang2022}. To address this, we propose using partition of unity (POU) networks integrated with PINNs to form a partition of unity physics-informed neural networks (POU-PINNs). A POU network divides the computational domain into localized subdomains, ensuring features are represented where they are most relevant \cite{Lee2021, Melenk1996}. This decomposition is achieved by employing POU functions that sum to one and are supported over distinct spatial regions \cite{Trask2021}. By combining the domain decomposition capabilities of partition of unity (POU) networks with the physics-informed framework of PINNs, POU-PINNs effectively address problems with significant spatial variability, such as those involving complex conductivity parameters. This hybrid approach enables precise solutions to PDEs by leveraging localized representations to capture distinct physical phenomena while narrowing the solution space \cite{Fan2023}. 

\subsection{Application on Thermal Ablation}
In the early NASA missions, blunt nose designs were used to evenly distribute airflow and heat transfer across the spacecraft's surface \cite{Amar2006}. To understand this process, ablation models were created to simulate surface heating and the breakdown of heat shield materials during re-entry \cite{Amar2022}. Based on the Darcy equation, these models included porous media flow dynamics to represent the evaporation and changes in the metal components within the heat shield's carbon layers during thermal ablation \cite{Kumar2016, Kotteda2019}.
\\
\\
Traditional methods for simulating thermal ablation used numerical techniques such as the finite element, finite volume, and finite difference methods aimed at solving partial differential equations (PDEs) \cite{Dec2010}. Thermal ablation involves several coupled models, including Darcy flow, which describes porosity dynamics, and material recession, which focuses on the behavior of carbon-based materials in thermal protection systems \cite{Mansour2024, Bychkov1993}. Additionally, the Richtmyer-Meshkov instability, driven by the Rayleigh-Taylor effect, occurs along the detached bow shock wave during spacecraft re-entry \cite{Zhou2021}.
To address the complexity of ablation physics, we developed a Partition of Unity Physics-Informed Neural Network (POU-PINN) solver grounded in Darcy flow principles. This solver leverages the partition of unity approach to decompose the domain into localized subdomains, enabling the accurate representation of spatially varying properties. Validated against analytical solutions from manufactured problems, the POU-PINN is implemented using TensorFlow/Keras \cite{Freno2021}. Combining physics-based models and data efficiently reduces the solution space, enhances convergence speed, and achieves results with machine-precision accuracy.

\subsection{Application on Ice-Sheet Modeling}
Our approach begins by using synthetically manufactured solutions designed to mimic the properties of Darcy flow, which are solved through PDE-based optimization. Manufactured solutions are simpler, allowing us to incrementally increase complexity until reaching a real-world PDE system with boundary conditions integrated into the partition of unity method presented in this paper. This approach also enables verification, as the true and exact solutions are always known within the manufactured solution framework \cite{Roache2002, Salari2000}.
\\
\\
In current studies, Darcy flow equations have been approximated using the finite element method (FEM), a state-of-the-art approach for modeling ice sheets and subsurface displacement. Sandia’s Albany code has validated FEM for over two decades, and the U.S. Department of Energy (DOE) is currently conducting experiments in Greenland and Antarctic ice shelves as a testbed for simulations performed using Albany \cite{Salinger2016, Tezaur2015, Hoffman2018}. The ultimate goal of this research is to achieve the performance level of Albany’s FEM implementation.
\\
\\
However, current state-of-the-art sensors provide only sparse and discrete measurements of critical parameters like viscosity, which impacts conductivity and displacement values. This paper focuses on synthetic scenarios to calculate conductivity and derive scalar fields from PDE solutions as a precursor to solving real PDE problems with boundary conditions. Analytical solutions, validated using Oberkampf's methodology, offer a baseline but fall short when compared to solving complex PDEs\cite{Oberkampf2002, Oberkampf2010}. Although extrapolation methods can partially predict future ice sheet displacement, many parameters are dynamic and unpredictable, often resulting in chaotic bifurcations and behaviors described in studies of chaotic systems \cite{Brunton2022}.
\\
\\
To address this unpredictability, we emphasize the importance of uncertainty quantification techniques \cite{Adams2014, Adams2020}. Sensitivity analyzes, polynomial chaos expansion, reinforcement learning, Bayesian and Markov approaches, and Monte Carlo sampling are crucial for tracking and managing uncertainty as it evolves over space and time \cite{Rodriguez2020, Paez2022, Kotteda2018}.
\\
\\
These approaches led to the discovery of qualitative physics in soft matter encounters in places where ice sheets and sub-surface flow exist, such as Greenland and the Antarctic ice shelves \cite{Cheng2024, Wang2022, Iwasaki2023}. In this paper, we refer to conductivity to quantify the movement of ice shelves within Greenland and the Antarctic ice shelves. Over space and time, they portray the ice sheet movement in Greenland and the Antarctic, which can be used as training sets for our new method to capture qualitative physics in ice sheets. We have taken an unsupervised approach to this problem since we do not need to manually classify the different physics for the fields, such as in traditional methods. The partition of the unity network can quantify these characteristics and apply them to subdomains as partitions, then calculate the scalar Darcy fluid flow field using the POU-PINNs approach.

\section{Problem Definition}
We have taken the forward design approach for the physics-informed neural network, where we give our partial differential equation and boundary conditions to compute the solution of our PDE and Boundary Conditions (BCs) of any rank, a vector field, or, in most of our demonstrated cases, a scalar field. First, we will describe the forward design of physics-informed neural networks.

\subsection{Physics-Informed Neural Networks}
The generic form of a steady-state equation that is formulated as a forward problem is the following:

\begin{align}
    \mathcal{N}[u, K] &= 0, \quad x \in \Omega
\end{align}
\begin{align}
    \mathcal{B}[u, K] &= 0, \quad x \in \partial \Omega
\end{align}
\\
Where $\mathcal{N}$ stands for the residual neural network of our PDE, $u(x)$ being the solution of the PDE, $K(x)$ a parameter and $\Omega$ the domain of interest. $\mathcal{B}$ stands for the boundary neural network that defines boundary conditions on the domain boundary $\partial \Omega$.    
\\
\\
In this work, we focus on a diffusion problem:
\\
\begin{align}
\begin{cases} 
\nabla \cdot (-K \nabla u) = f \quad \text{in } \Omega \\
-K \nabla u \cdot \hat{n} = g_N \quad \text{on } \Gamma_N \\
u = g_D \quad \text{on } \Gamma_D
\end{cases}
\end{align}
\\
\\
where $\partial \Omega$ is partitioned into $\Gamma_N$ and $\Gamma_D$, where we prescribe Neumann and Dirichlet conditions, respectively. $K$ is the diffusivity coefficient, $f$ the forcing term.
\\
\\
We are interested in problems with different diffusivity values in different parts of the domain. In general, if we partition the domain $\Omega$ into $M$ domains $\Omega_i$, $i = 1, \ldots, M$, then $K(x) = K_i$ for $x \in \Omega_i$.
\\
\\
In this case, we must require the continuity of the normal fluxes at the interfaces, $\Gamma_{ij}$ of domains $\Omega_i$ and $\Omega_j$:
\\
\\
\begin{align}
K_i \nabla u \cdot \vec{n}_{ij} = K_j \nabla u \cdot \vec{n}_{ij} \quad \text{on } \Gamma_{ij}
\end{align}
\\
\\
Here $\vec{n}_{ij}$ is the normal of $\Gamma_{ij}$ outward of $\Omega_i$.

\section{Methods}

\subsection{Physics-Informed Neural Networks}

The physics-informed neural network learns from known priors depending on the amount of information given. Our scenario contains the partial differential governing equations and the boundary conditions for a spatial-dependent problem, as shown in Fig. 1. Then, we derive our comprehensive loss functional being the residual for all components of the problem:
\begin{align}
\mathcal{L}(w) = \mathcal{L}_{\text{PDE}}(w) + \mathcal{L}_{\text{BC}}(w)
\end{align}
\\ \\
The PDE component loss functional:
\begin{align}
\mathcal{L}_{\text{PDE}}(w) = \frac{1}{N} \sum_{i=1}^{N} \left| \mathcal{N}_x \big(u(x_{\text{PDE}}^i; w)\big) \right|^2
\end{align}
The BC component loss functional:
\begin{align}
\mathcal{L}_{\text{BC}}(w) = \frac{1}{N} \sum_{i=1}^{N} \left| \mathcal{B}_x \big[u(x_{\text{BC}}^i; w)\big] \right|^2
\end{align}

\begin{figure}[h!]
    \centering
    \includegraphics[width=1.0\textwidth]{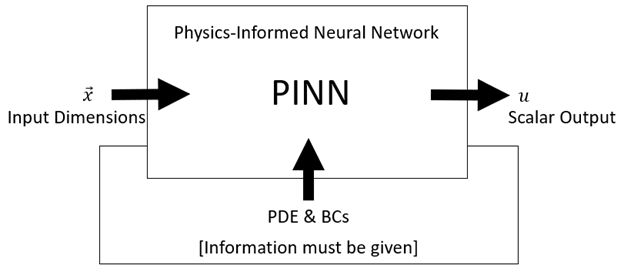} 
    \caption*{\textbf{Fig.1 Physics-informed neural network:} In our cases, we consider only the input dimensions, the partial differential equation, and boundary conditions as inputs, and the solution of the physics-informed neural network as a scalar output for the forward design of the neural network.}
    \label{fig:pinn}
\end{figure}

\subsection{Partition of Unity Networks}

We use the partition of unity networks to compute the conductivity. The conductivity $K_{\text{POU}}$ is computed by summing over all points while computing the $\varphi$ functions and $c$ coefficients as in Fig.~2. We use a monotonic exponential function to make the output of the conductivity always positive using the following function:

\begin{align}
K_{\text{POU}}(x) = \sum_{i=1}^{N} \varphi_i(x) e^{c_i}
\end{align}
\\ 
We apply the following conditions to satisfy unity [7].
\begin{align}
\sum_i \varphi_i(x) = 1
\end{align}
This is done using the code's SoftMax activation function. SoftMax constrains the values from zero to one and ensures that all functions add up to one in the whole space, as in Fig. 3. We minimize the difference between the neural network calculations and the training data.
\begin{align}
\arg\min_{\zeta, c} \sum_{i \in D} \left| \sum_{i=1}^{N} \varphi_i(x_i, \zeta) e^{c_i} - K_i \right|^2
\end{align}
\\
Where the partition functions are a neural network representation.
\begin{align}
\varphi_i(x; \zeta) = \big[\mathcal{N}\mathcal{N}(x; \zeta)\big]_i
\end{align}

\begin{figure}[h!]
    \centering
    \includegraphics[width=1.0\textwidth]{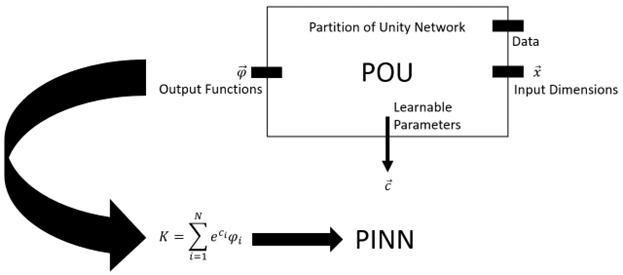} 
    \caption*{\textbf{Fig.2 Partition of Unity Networks Leading to PINN:} This figure shows our approach by using the data and the input dimensions as a vector of inputs $(x \text{ and } y)$. Then, the partition of unity network outputs as it learns the $c$ learnable parameters and $\varphi$ functions are used to compute the conductivity using the following equation, which is then inserted into the forward-designed physics-informed neural network.}
    \label{fig:pinn}
\end{figure}

\begin{figure}[h!]
    \centering
    \includegraphics[width=0.7\textwidth]{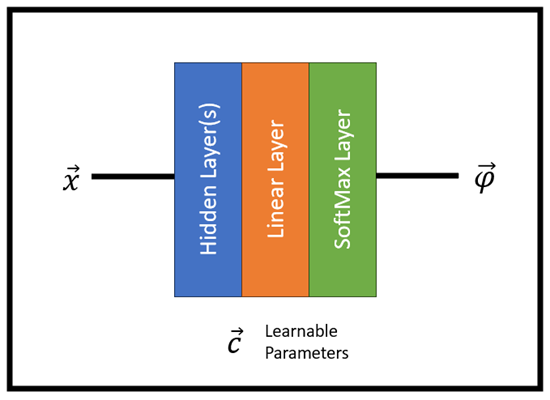} 
    \caption*{\textbf{Fig.3 Partition of Unity Networks Architecture:} The architecture of the partition of the unity network consists of inputs, hidden layer(s), a linear activation function layer, and a softmax activation function layer, leading to the output functions $\vec{\varphi}$.}
    \label{fig:pinn}
\end{figure}

\subsection{Partition of Unity Physics-Informed Neural Networks}

Partition of unity physics-informed neural networks (pou-pinns) can capture the partitions of space that represent a localized conductivity parameter that is localized for the solution of the PDE on that partition. The PDE governing the porous media modeling the physics is the Darcy flow equation that is parameterized by a conductivity parameter, where the conductivity is different throughout the solution—knowing that the partitions could capture the qualitative physics and find the conductivity parameter based on images. We have taken the unsupervised approach where space decompositions could be discovered and used on the Darcy equation's parametrized form, where the physics' localized solution is found as in Fig. 4.

\begin{figure}[h!]
    \centering
    \includegraphics[width=0.9\textwidth]{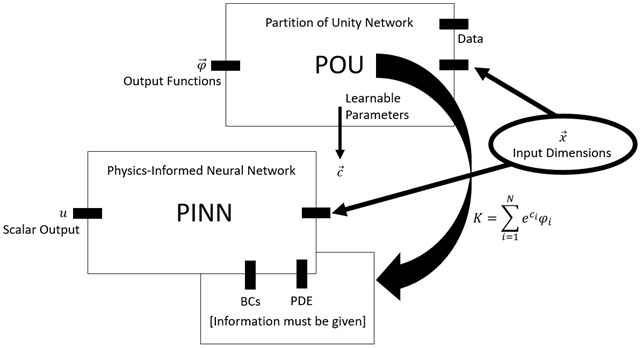} 
    \caption*{\textbf{Fig.4. Partition of Unity Network Physics-Informed Neural Networks:} Here, we show a schematic representation of the partition of unity networks and the physics-informed neural networks and how they are connected to create a partition of unity physics-informed neural networks.}
    \label{fig:pinn}
\end{figure}

\section{Results and Discussion}

\subsection{Physics-Informed Neural Networks - Example 1}

In this case, in Fig. 5, 6, and 7, we consider the problem (Eq.3 set) on the unit square 
\(\Omega = [0,1]^2\) and homogeneous Dirichlet conditions 
\((\Gamma_D = \partial \Omega, \Gamma_N = \emptyset, g_D = 0):\)

\begin{align}
\begin{cases} 
\nabla \cdot \left( -K \nabla u \right) = f & \quad \text{in } \Omega, \\
u = 0 & \quad \text{on } \partial \Omega,
\end{cases}
\end{align}
\\
For this scenario, our true solution is the following:
\begin{align}
u_{\text{true}} = \sin(2\pi x) \cdot \sin(2\pi y)
\end{align}
\\
Hence, assuming (eq.12) holds, the forcing term is: 
\begin{align}
f(x, y) = -8 \pi^2 K \sin(2\pi x) \cdot \sin(2\pi y), \quad (x, y) \in \Omega_i
\end{align}
Where \( K = 1 \) for this case but can be indexed, \( K_i \) with variable conductivity.
This is the solution to our PDE:
\begin{figure}[h!]
    \begin{subfigure}[b]{0.45\textwidth}
        \centering
        \includegraphics[width=\textwidth]{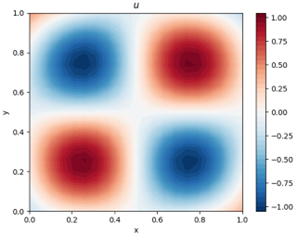} 
        \caption{2D PDE Solution}
        \label{fig:2D PDE Solution}
    \end{subfigure}
    \hfill
    \begin{subfigure}[b]{0.45\textwidth}
        \centering
        \includegraphics[width=\textwidth]{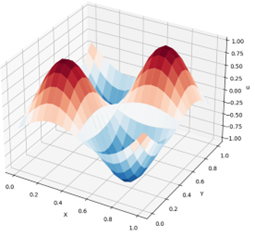} 
        \caption{3D PDE Solution}
        \label{fig:3D PDE Solution}
    \end{subfigure}
    \caption*{\textbf{Fig.5 Scalar field PDE solution:} Scalar field solution is computed using the forward design of PINNs. The solution demonstrates four subdomains with equal geometric characteristics and two pairs with equal solutions on different subdomains. Then, the three-dimensional representation of our solution is on the right.}
    \label{fig:combined}
\end{figure}

\begin{figure}[h!]
    \centering
    \includegraphics[width=0.8\textwidth]{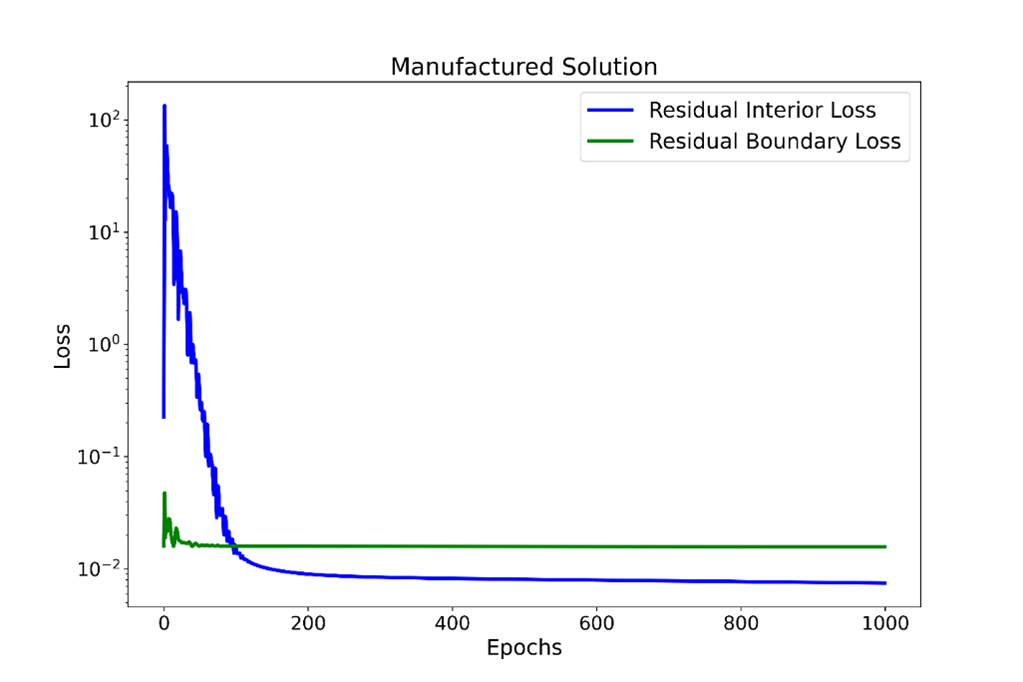} 
    \caption*{\textbf{Fig.6 Manufactured Solution Residuals:} Our method found the domain's solution on 1000 epochs. Since the boundaries are already satisfied by their initialization being zeros, it does not need to learn the solution of the boundaries.}
    \label{fig:pinn}
\end{figure}

\begin{figure}[h!]
    \centering
    \includegraphics[width=0.7\textwidth]{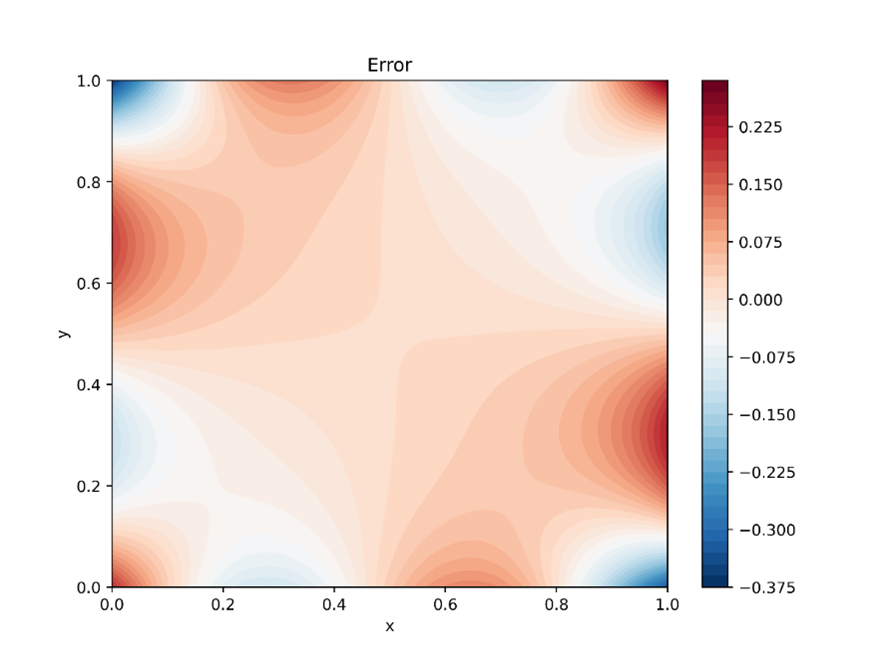} 
    \caption*{\textbf{Fig.7 Manufactured solution against true solution error contour:} This contour demonstrates the agreement between the manufactured solution computed using PINNs against the true solution of the example 1 problem.}
    \label{fig:pinn}
\end{figure}

\subsection{Physics-Informed Neural Networks - Example 2}

In this case, in Fig. 8, 9, 10, and 11, we consider the problem (Eq. 3 set) on the unit square 
\(\Omega = [0,1]^2\) and homogeneous Dirichlet on \(x\) and Neumann conditions on \(y\) 
\((\Gamma_N = \partial \Omega, \Gamma_D = \partial \Omega, g_N = 0, g_D = 0):\)

\begin{align}
\begin{cases} 
\nabla \cdot (-K \nabla u) = f \quad \text{in } \Omega \\
-K \nabla u \cdot \hat{n} = g_N \quad \text{on } \Gamma_N \\
u = g_D \quad \text{on } \Gamma_D
\end{cases}
\end{align}
\\
For this scenario, our true solution is the following:
\begin{align}
u_{\text{true}} = \sin(m \pi x)
\end{align}
Hence, assuming (eq.15) holds, the forcing term is:

\begin{align}
f(x, y) = \left(m^2 \pi^2\right) \sin(m \pi x), \quad (x, y) \in \Omega_i    
\end{align}
Where m = 2 and K = 1.

\begin{figure}[h!]
    \begin{subfigure}[b]{0.45\textwidth}
        \centering
        \includegraphics[width=\textwidth]{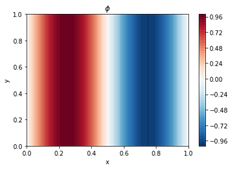} 
        \caption{2D PDE Solution}
        \label{fig:2D PDE Solution}
    \end{subfigure}
    \hfill
    \begin{subfigure}[b]{0.45\textwidth}
        \centering
        \includegraphics[width=\textwidth]{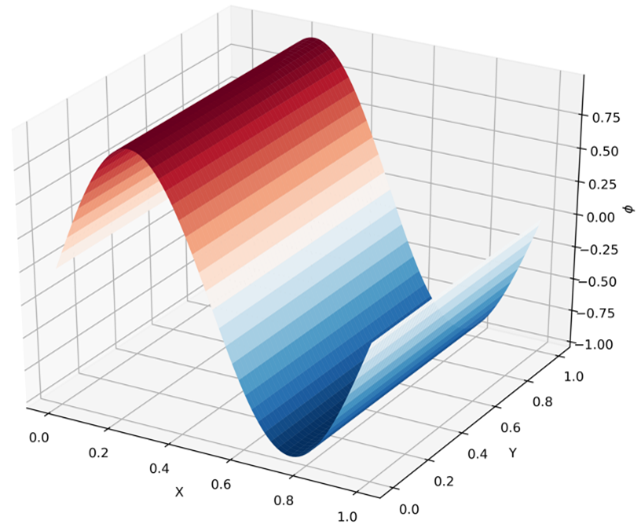} 
        \caption{3D PDE Solution}
        \label{fig:3D PDE Solution}
    \end{subfigure}
    \caption*{\textbf{Fig.8 Scalar field PDE solution:} Scalar field solution is computed using the forward design of PINNs. The solution demonstrates two subdomains with equal geometric characteristics. Then, the three-dimensional representation of our solution is on the right.}
    \label{fig:combined}
\end{figure}

\begin{figure}[h!]
    \centering
    \includegraphics[width=0.6\textwidth]{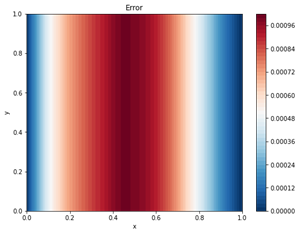} 
    \caption*{\textbf{Fig.9 Manufactured solution against true solution error contour:} This contour demonstrates the agreement between the manufactured solution computed using PINNs against the true solution of the example 2 problem.}
    \label{fig:pinn}
\end{figure}

\vspace{18em}

\begin{figure}[h!]
    \centering
    \includegraphics[width=1.0\textwidth]{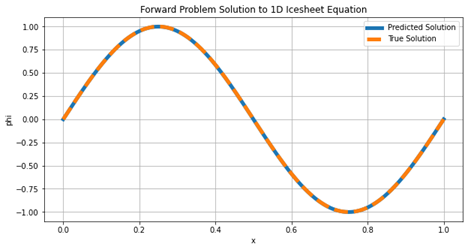} 
    \caption*{\textbf{Fig.10 True Solution Comparison:} The true solution agrees with the predicted solution.}
    \label{fig:pinn}
\end{figure}

\begin{figure}[h!]
    \centering
    \includegraphics[width=0.8\textwidth]{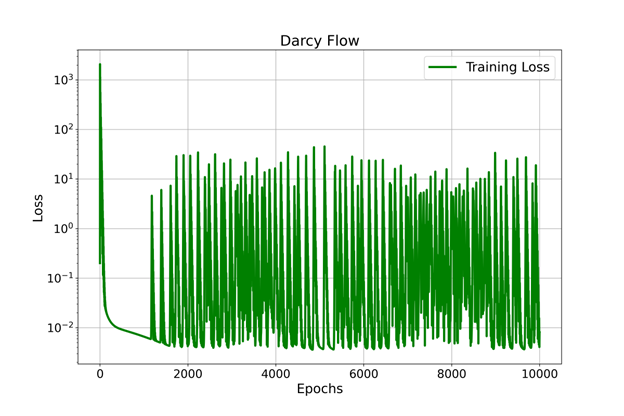} 
    \captionsetup{justification=raggedright, singlelinecheck=false} 
    \caption*{\textbf{Fig.11 Two-Partitions Darcy Flow Residuals:} Loss Functional of Darcy Flow.}
    \label{fig:figure11}
\end{figure}

\vspace{66em}

\subsection{Partition of Unity - Example 1}

We have two partitions in space that are governed by the following expression:

\begin{align}
K(x, y) = 1 \cdot \left( \frac{1 - \text{sign}(x + y - 1)}{2} \right) + 10 \cdot \left( \frac{1 + \text{sign}(x + y - 1)}{2} \right)
\end{align}
\\
In the scenario depicted in Fig. 12, we utilized the partition of unity networks to compute the functions \(\phi\), which represent spatial partitions and correspond to distinct conductivities influencing the physics of the PDE solution. To construct \(K_{\text{POU}}\), each function \(\phi\) is multiplied by its respective learned coefficient within the network.
\\
\\
To achieve convergence for this problem, the model was trained for 1000 epochs using the Adam optimizer with a learning rate of 0.0005 \cite{Kingma2014}. The architecture for this 2D spatial problem included a single hidden layer with 40 neurons, utilizing a hyperbolic tangent activation function. The linear layer consisted of 40 neurons with a linear activation function, followed by an output layer with 40 neurons and a SoftMax activation function \cite{Nwankpa2018}.
\\
\\
We set the random seed to 12345 for reproducibility in the Python NumPy and TensorFlow libraries \cite{Madhyastha2019}. We found it highly advantageous to generate training points using the analytic expressions during training, as this approach allows the network to learn more effectively. However, this luxury is often unavailable in real-world problems due to the scarcity of in-situ measurements and the limited number of simulations.
\\
\\
The partition of unity network proved an excellent choice for our application, as it excels at computing composite parameterizations. This capability was crucial for accurately determining conductivity within the partition of unity physics-informed neural network framework.

\begin{figure}[h!]
    \begin{subfigure}[b]{0.45\textwidth}
        \centering
        \includegraphics[width=\textwidth]{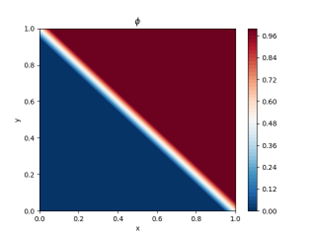} 
        \caption{2D Partition}
        \label{fig:2D Partition}
    \end{subfigure}
    \hfill
    \begin{subfigure}[b]{0.45\textwidth}
        \centering
        \includegraphics[width=\textwidth]{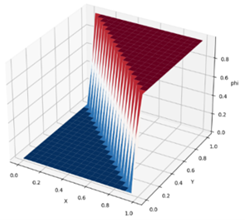} 
        \caption{3D Partition}
        \label{fig:3D Partition}
    \end{subfigure}
    \caption*{\textbf{Fig.12. Partition of Unity Networks (Two Partitions Triangles Scenario):} In this example, we have computed the solution of the two triangles problem using the partition of unity networks with two partitions where we show the two-dimensional and three-dimensional representations.}
    \label{fig:combined}
\end{figure}

\subsection{Partition of Unity - Example 2}

In Fig. 13, we have four partitions in space that are governed by the following expression:

\begin{align}
K(x, y) = 1 \cdot \left( \frac{1 - \text{sign}(x + y - 1)}{2} \right) + 10 \cdot \left( \frac{1 + \text{sign}(x + y - 1)}{2} \right)
\end{align}
\\
This expression is similar to the previous one; however, instead of identifying two spatial partitions with two different conductivities, we aimed to identify four spatial partitions, each with a unique conductivity. To achieve convergence, the model was trained for 600 epochs using the Adam optimizer with a learning rate of 0.00025.
\\
\\
The architecture for this 2D spatial problem included a single hidden layer with 40 neurons and a hyperbolic tangent activation function. The linear layer also consisted of 40 neurons with a linear activation function, followed by an output layer with 40 neurons and a SoftMax activation function. To ensure stable convergence, we applied L2 regularization with a value of 0.000001, manually calibrated \cite{Cortes2012, VanLaarhoven2017}.
\\
\\
The random seed was set to 12345 for reproducibility in both the Python NumPy and TensorFlow libraries. This ensured consistent results across experiments.

\begin{figure}[h!]
    \centering
    \begin{subfigure}[b]{0.45\textwidth}
        \centering
        \includegraphics[width=\textwidth]{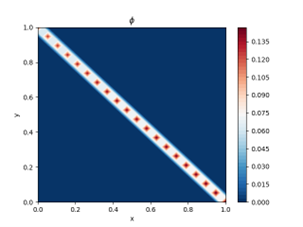} 
        \caption{2D Partition}
        \label{fig:image1}
    \end{subfigure}
    \hfill
    \begin{subfigure}[b]{0.45\textwidth}
        \centering
        \includegraphics[width=\textwidth]{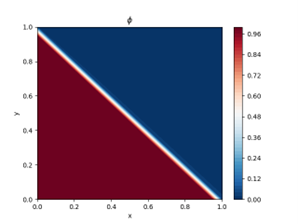} 
        \caption{2D Partition}
        \label{fig:image2}
    \end{subfigure}
    \vspace{1em}
    \begin{subfigure}[b]{0.45\textwidth}
        \centering
        \includegraphics[width=\textwidth]{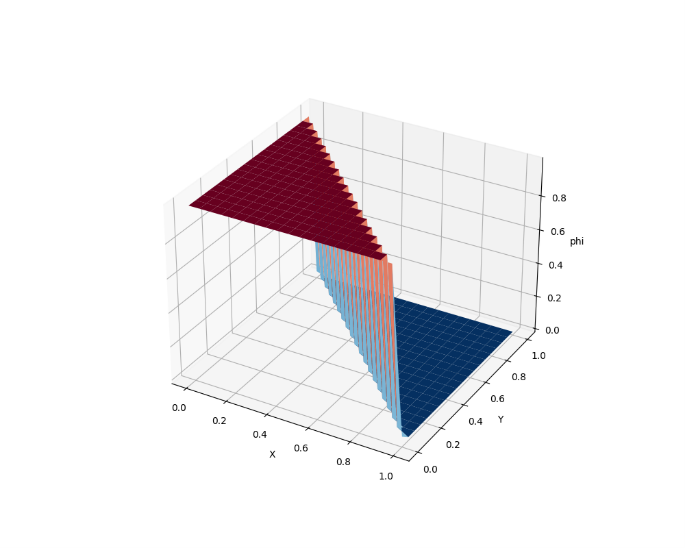} 
        \caption{3D Partition}
        \label{fig:image3}
    \end{subfigure}
    \hfill
    \begin{subfigure}[b]{0.45\textwidth}
        \centering
        \includegraphics[width=\textwidth]{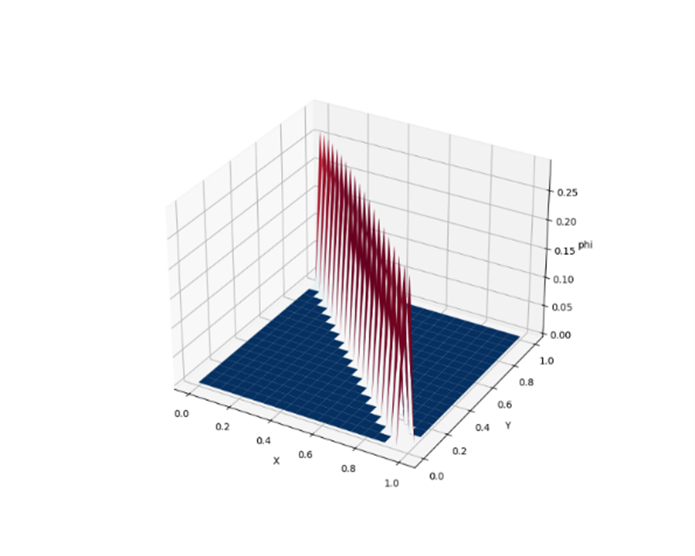} 
        \caption{3D Partition}
        \label{fig:image4}
    \end{subfigure}
    \captionsetup{justification=raggedright, singlelinecheck=false} 
    \renewcommand{\thefigure}{Custom Figure Label} 
    \caption*{\textbf{Fig.13 Partition of Unity Networks (Four Partitions Triangles Scenario):} In this example, we have computed the solution of the two triangles problem using the partition of unity networks with four partitions where we show the two-dimensional and three-dimensional representations.}
    \label{fig:matrix}
\end{figure}

\subsection{Partition of Unity - Example 3}

In Fig. 14, we trained the model for 10,000 epochs using the Adam optimizer with a learning rate of 0.000125 to achieve convergence. The architecture for this 2D spatial problem consisted of four hidden layers, each with 40 neurons and a hyperbolic tangent activation function. The output layer also had 40 neurons and utilized a SoftMax activation function.
\\
\\
To ensure stable convergence, we applied L2 regularization with a manually calibrated value of 0.000001. We set the random seed to 12345 for reproducibility in both the Python NumPy and TensorFlow libraries. Furthermore, we used the Glorot uniform initialization of weights with the same seed value \cite{Glorot2010}.
\\
\\
We employ transfer learning to enhance accuracy, taking advantage of the results of seven previous training runs. This approach enabled us to achieve machine-level precision while solving for four distinct partitions, each with unique conductivity values.

\begin{figure}[h!]
    \centering
    \begin{subfigure}[b]{0.45\textwidth} 
        \centering
        \includegraphics[width=\textwidth]{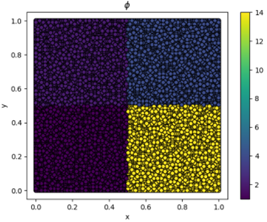} 
        \caption{}
        \label{fig:combined_top}
    \end{subfigure}
    \vspace{1em} 
    \begin{subfigure}[b]{0.45\textwidth}
        \centering
        \includegraphics[width=\textwidth]{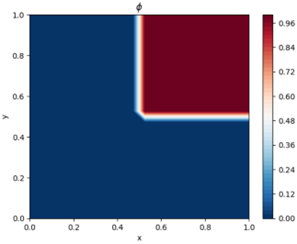} 
        \caption{}
        \label{fig:bottom_left}
    \end{subfigure}
    \hfill
    \begin{subfigure}[b]{0.45\textwidth}
        \centering
        \includegraphics[width=\textwidth]{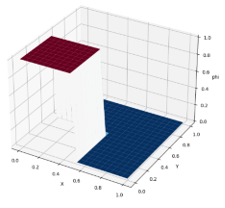} 
        \caption{}
        \label{fig:bottom_right}
    \end{subfigure}
    \captionsetup{justification=raggedright, singlelinecheck=false} 
    \caption*{\textbf{Fig.14. Partition of Unity Networks (Four Partitions Shuffled Boxes Scenario):} 
    In this example, we have computed the solution of the four-shuffled boxes problem using the partition of unity networks with four partitions, where we show the two-dimensional and three-dimensional representations.}
    \label{fig:partition_unity}
\end{figure}

\subsection{Partition of Unity - Example 4}

In the case shown in Fig. 15, values were appended to the boundary points to ensure accurate solution computation using the Partition of Unity Networks. The number of partitions was set to 4, representing four distinct conductivity values.
\\ \\
To achieve convergence, the model was trained for 3000 epochs using the Adam optimizer with a learning rate of 0.000125. The network architecture for this 2D spatial problem included two hidden layers, each with 40 neurons and a hyperbolic tangent activation function, followed by an output layer with 40 neurons and a SoftMax activation function.
\\ \\
To ensure stability, we applied L2 regularization with a value of 0.000001 and manually fine-tuned for this problem. We set the random seed to 12345 for reproducibility in both the Python NumPy and TensorFlow libraries. This setup facilitated consistent and reliable results across runs.

\begin{figure}[h!]
    \centering
    \begin{subfigure}[b]{0.45\textwidth} 
        \centering
        \includegraphics[width=\textwidth]{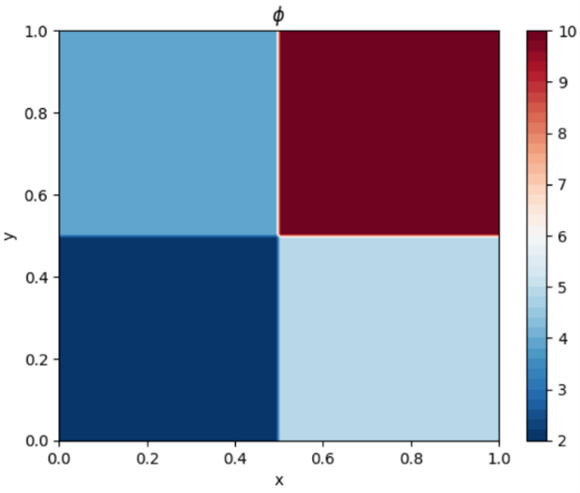} 
        \caption{}
        \label{fig:combined_top}
    \end{subfigure}
    \vspace{1em} 
    \begin{subfigure}[b]{0.45\textwidth}
        \centering
        \includegraphics[width=\textwidth]{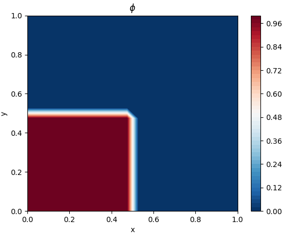} 
        \caption{}
        \label{fig:bottom_left}
    \end{subfigure}
    \hfill
    \begin{subfigure}[b]{0.45\textwidth}
        \centering
        \includegraphics[width=\textwidth]{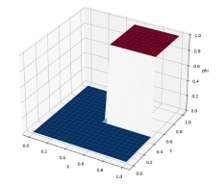} 
        \caption{}
        \label{fig:bottom_right}
    \end{subfigure}
    \captionsetup{justification=raggedright, singlelinecheck=false} 
    \caption*{\textbf{Fig.15 Partition of Unity Networks (Four Partitions Boxes Scenario):} In this example, we computed the solution of the four boxes problem using the partition of unity networks with four partitions, where we show the two-dimensional and three-dimensional representations.}
    \label{fig:partition_unity}
\end{figure}

\subsection{Partition of Unity - Example 5}

In Fig. 16, we have two partitions in space that are governed by the following expression:

\begin{align}
K(x, y) = 1 \cdot \left( \frac{1 - \text{sign}(x - 0.5)}{2} \right) + 4 \cdot \left( \frac{1 + \text{sign}(x - 0.5)}{2} \right)
\end{align}
\\
To achieve convergence for this problem, the model was trained for 100 epochs using the Adam optimizer with a learning rate of 0.00025. The network architecture for this 2D spatial problem included a single hidden layer with 40 neurons and a hyperbolic tangent activation function, followed by a linear activation layer and an output layer with 40 neurons utilizing a SoftMax activation function.
\\ \\
To ensure stable convergence, L2 regularization with a value of 0.000001 was applied and manually calibrated for optimal performance. The random seed was set to 12345 for reproducibility in both the Python NumPy and TensorFlow libraries. This configuration enabled consistent and accurate results across training runs.

\begin{figure}[h!]
    \centering
    \begin{subfigure}[b]{0.30\textwidth} 
        \centering
        \includegraphics[width=\textwidth]{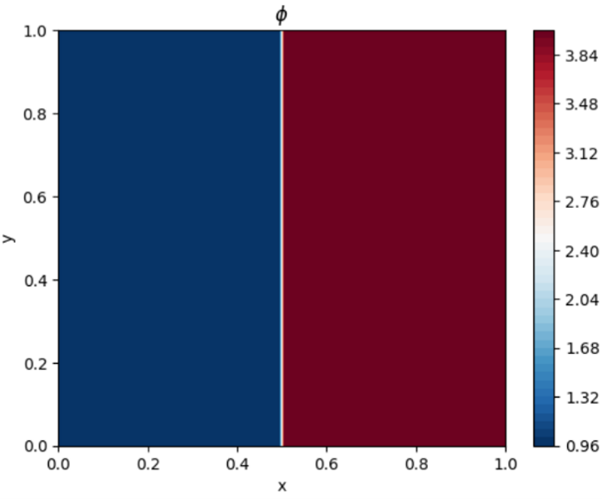} 
        \caption{}
        \label{fig:combined_top}
    \end{subfigure}
    \vspace{1em} 
    \begin{subfigure}[b]{0.31\textwidth}
        \centering
        \includegraphics[width=\textwidth]{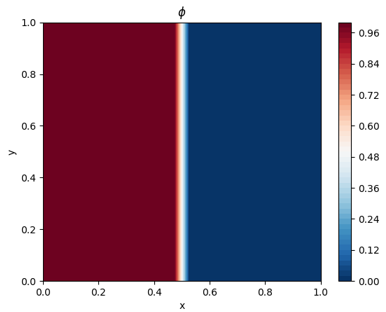} 
        \caption{}
        \label{fig:bottom_left}
    \end{subfigure}
    \hfill
    \begin{subfigure}[b]{0.30\textwidth}
        \centering
        \includegraphics[width=\textwidth]{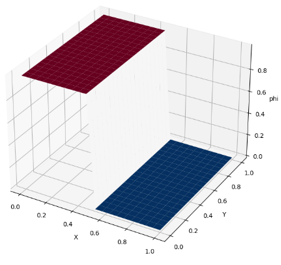} 
        \caption{}
        \label{fig:bottom_right}
    \end{subfigure}
    \captionsetup{justification=raggedright, singlelinecheck=false} 
    \caption*{\textbf{Fig.16 Partition of Unity Networks (Two Partitions Strips Scenario):} In this example, we have computed the solution of the two-strip problem using the Partition of Unity Networks with two partitions, where we show the two-dimensional and three-dimensional representations.}
    \label{fig:partition_unity}
\end{figure}

\subsection{Partition of Unity - Example 6}

In Fig. 17. problem, we have two partitions in space that are governed by the following expression:

\begin{align}
K(x, y) = 4 \cdot \left( \frac{1 - \text{sign}(x - 0.5)}{2} \right) + 1 \cdot \left( \frac{1 + \text{sign}(x - 0.5)}{2} \right)
\end{align}
\\
When we computed the solution of the Partition of Unity Networks, we selected the number of partitions equal to 2. In this case, we just inverted the previous case along the y-axis.

\begin{figure}[h!]
    \centering
    \begin{subfigure}[b]{0.30\textwidth} 
        \centering
        \includegraphics[width=\textwidth]{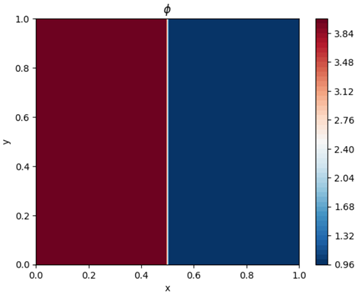} 
        \caption{}
        \label{fig:combined_top}
    \end{subfigure}
    \vspace{1em} 
    \begin{subfigure}[b]{0.31\textwidth}
        \centering
        \includegraphics[width=\textwidth]{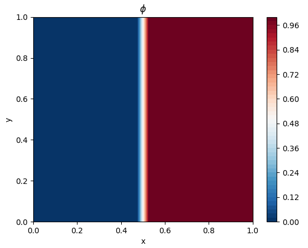} 
        \caption{}
        \label{fig:bottom_left}
    \end{subfigure}
    \hfill
    \begin{subfigure}[b]{0.30\textwidth}
        \centering
        \includegraphics[width=\textwidth]{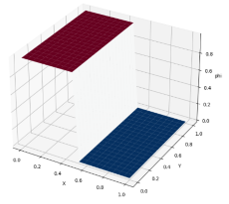} 
        \caption{}
        \label{fig:bottom_right}
    \end{subfigure}
    \captionsetup{justification=raggedright, singlelinecheck=false} 
    \caption*{\textbf{Fig.17 Partition of Unity Networks (Two Partitions Inverted Strips Scenario):} In this example, we have computed the solution of the two-strip inverted problem using the Partition of Unity Networks with two partitions, where we show the two-dimensional and three-dimensional representation.}
    \label{fig:partition_unity}
\end{figure}

\section{Partition of Unity Physics-Informed Neural Networks (POU-PINNs)}

\subsection{POU-PINNs - Example 1}

In partition of unity physics-informed neural networks, we use the partition of unity network to compute the conductivity of the porous media physics as an input for the physics-informed neural network to compute u: the scalar output field; the solution to the PDE in the physics-informed neural network, where the solution represents how pressure impacts the movement of the fluid.
\\ \\
Fig. 18, 19, 20, 21, 22, and 23 show Partition of Unity Physics-Informed Neural Networks.
\begin{figure}[h!]
    \centering
    \includegraphics[width=0.40\textwidth]{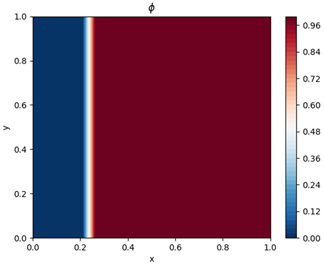} 
    \caption*{\textbf{Fig.18 POU-PINN - (Partition of Unity) Example 1:} Here is the output representation of one of the partitions of unity networks combined with physics-informed neural networks as a representation of that partition.}
    \label{fig:pinn}
\end{figure}
\\ \\
To achieve convergence for this problem, the model was trained for 6000 epochs using the Adam optimizer with a learning rate of 0.001.
\\ \\
For the Physics-Informed Neural Network (PINN), the architecture consisted of four hidden layers with 40 neurons each, utilizing a hyperbolic tangent activation function followed by a linear activation layer. For the Partition of Unity (POU) component, the architecture also featured four hidden layers with 40 neurons and a hyperbolic tangent activation function, with a linear activation layer and a SoftMax activation function for the output layer.
\\ \\
To ensure stable convergence, L2 regularization with a value of 0.0001 was applied and manually fine-tuned for this specific problem. The random seed was set to 12345 for reproducibility in the Python NumPy and TensorFlow/Keras libraries. This setup ensured consistent and reliable results across training runs.

\begin{figure}[h!]
    \begin{subfigure}[b]{0.45\textwidth}
        \centering
        \includegraphics[width=\textwidth]{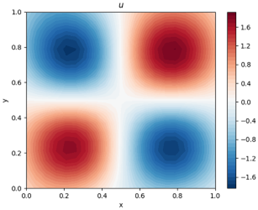} 
        \caption{2D PINN Contour}
        \label{fig:2D Partition}
    \end{subfigure}
    \hfill
    \begin{subfigure}[b]{0.45\textwidth}
        \centering
        \includegraphics[width=\textwidth]{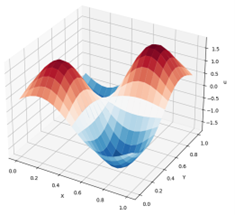} 
        \caption{3D PINN Contour}
        \label{fig:3D Partition}
    \end{subfigure}
    \captionsetup{justification=raggedright, singlelinecheck=false}
    \caption*{\textbf{Fig.19 POU-PINN - (Physics-Informed Neural Network) Example 1:} This scalar field represents the solution of the physics-informed neural networks when the partition of the unity physics-informed neural network is applied and simulated for this problem.}
    \label{fig:combined}
\end{figure}

\begin{figure}[h!]
    \centering
    \includegraphics[width=0.7\textwidth]{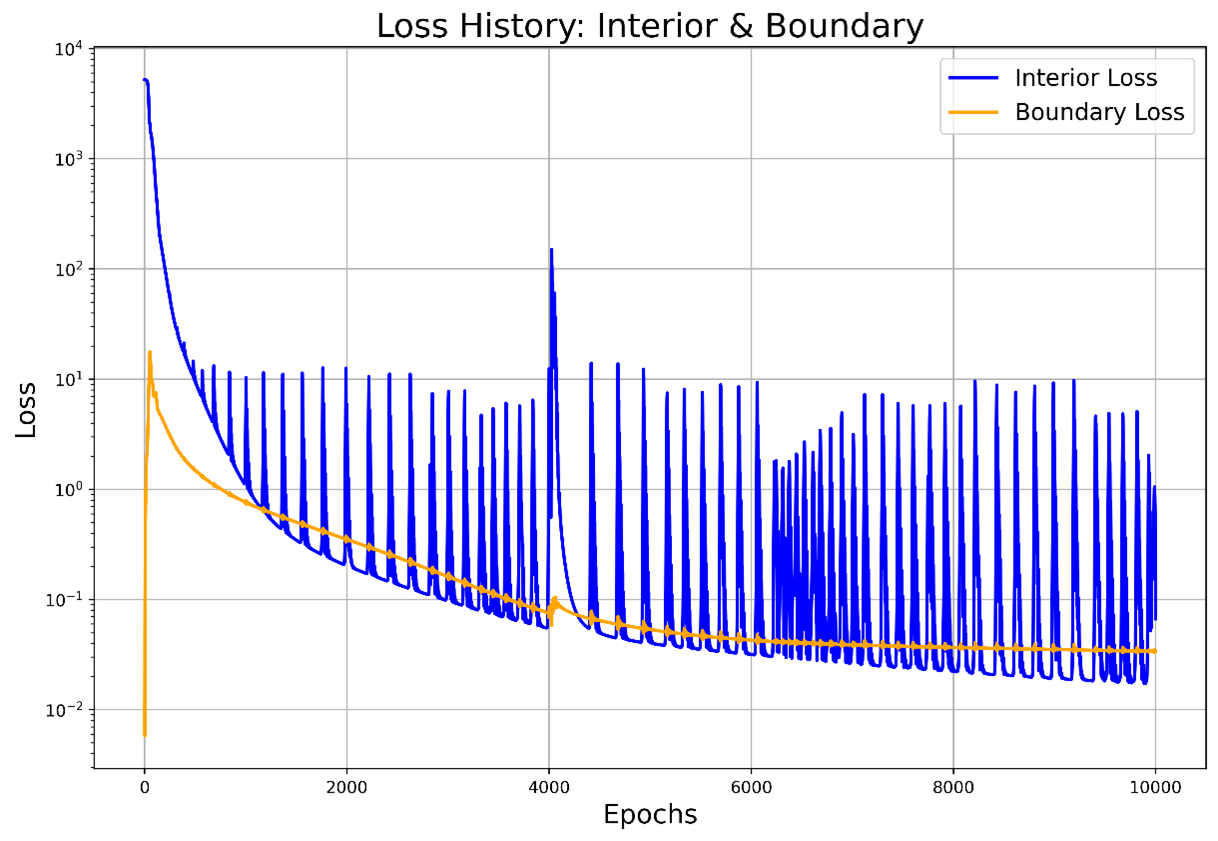} 
    \captionsetup{justification=raggedright, singlelinecheck=false}
    \caption*{\textbf{Fig.20 POU-PINN - Residuals Example 1:} The model did not maintain stability due to partition discontinuity, but it led to the correct solution.}
    \label{fig:figure20}
\end{figure}

\FloatBarrier 

\subsection{POU-PINNs - Example 2}

\begin{figure}[h!]
    \centering
    \includegraphics[width=0.4\textwidth]{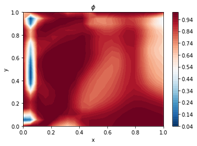} 
    \captionsetup{justification=raggedright, singlelinecheck=false}
    \caption*{\textbf{Fig.21 POU-PINN - (Partition of Unity) Example 2:} Here is the output representation of one of the partitions of unity networks combined with physics-informed neural networks as a representation of that partition.}
    \label{fig:figure21}
\end{figure}

\begin{figure}[h!]
    \begin{subfigure}[b]{0.45\textwidth}
        \centering
        \includegraphics[width=\textwidth]{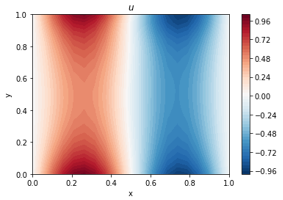} 
        \caption{2D PINN Contour}
        \label{fig:2D Partition}
    \end{subfigure}
    \hfill
    \begin{subfigure}[b]{0.45\textwidth}
        \centering
        \includegraphics[width=\textwidth]{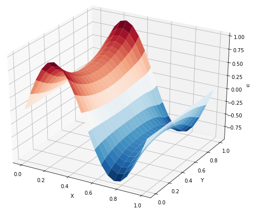} 
        \caption{3D PINN Contour}
        \label{fig:3D Partition}
    \end{subfigure}
    \captionsetup{justification=raggedright, singlelinecheck=false}
    \caption*{\textbf{Fig.22 POU-PINN (Physics-Informed Neural Network) Example 2:} This scalar field represents the solution of the physics-informed neural networks when the partition of the unity physics-informed neural network is applied and simulated for this problem.}
    \label{fig:combined}
\end{figure}

\begin{figure}[h!]
    \centering
    \includegraphics[width=0.7\textwidth]{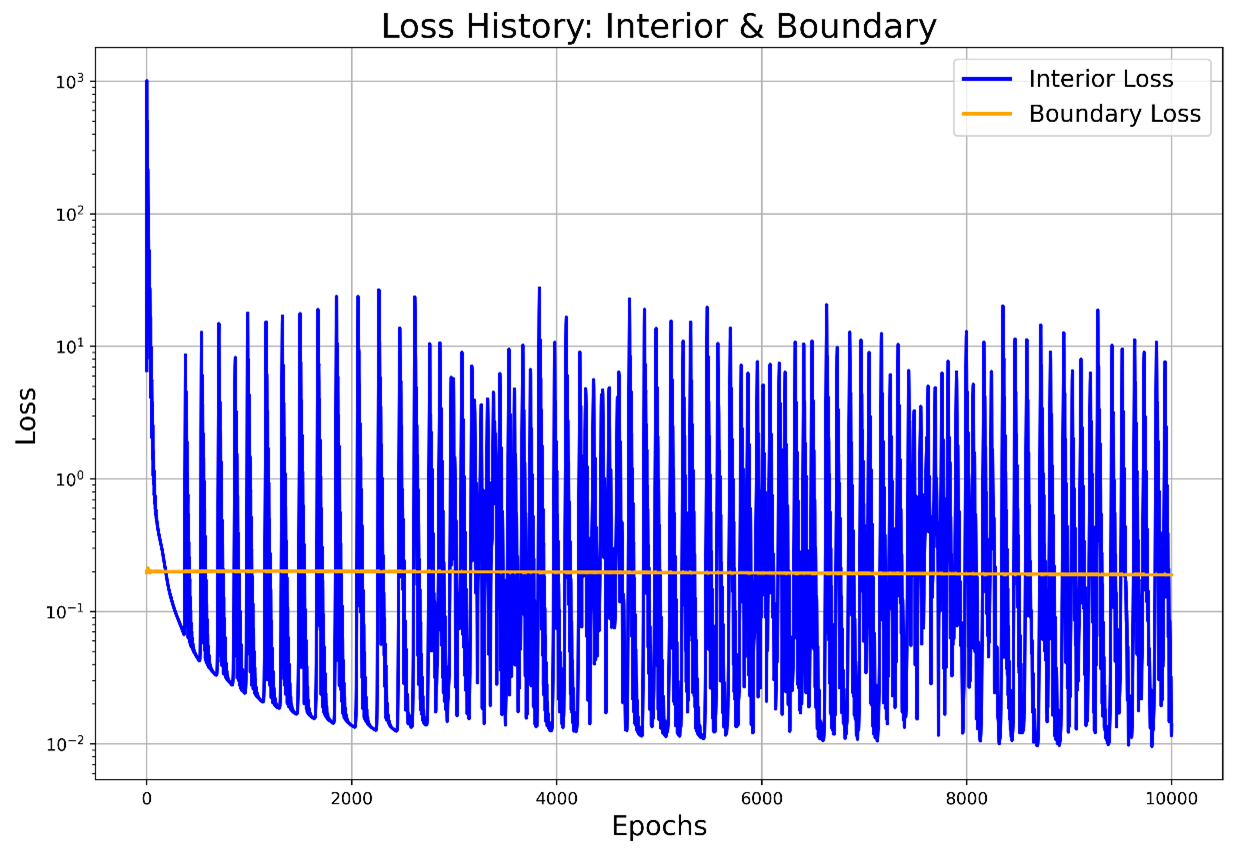} 
    \captionsetup{justification=raggedright, singlelinecheck=false}
    \caption*{\textbf{Fig.23 POU-PINN Residuals Example 2:} The model did not maintain stability due to partition discontinuity but led to the correct solution.}
    \label{fig:figure20}
\end{figure}

\FloatBarrier

\section{Conclusion}

Combining the partition of unity networks with physics-informed neural networks, we have developed a robust computational framework called Partition of Unity Physics-Informed Neural Networks (POU-PINNs). This innovative model integrates the strengths of partition of unity methods for space decomposition with the physics-based learning capabilities of neural networks, enabling efficient modeling of complex physical systems characterized by localized variations.
\\ \\
The POU-PINN approach involves partitioning the computational domain into subdomains, each represented by unique localized functions that sum to unity, ensuring smooth transitions across subdomains. By incorporating this spatial decomposition into physics-informed neural networks, the model effectively handles forward and inverse problems involving partial differential equations (PDEs), particularly in scenarios with significant spatial variability, such as porous media and ice-sheet dynamics.
\\ \\
In this work, we utilized POU-PINNs to solve a forward design problem, leveraging the localized framework to optimize PDE solutions. Additionally, we parameterized the equations using conductivity parameters and tackled an inverse problem, showcasing the model's capability to infer system properties from observed data. Applying thermal ablation within porous media demonstrated the framework's ability to model complex heat transfer phenomena. At the same time, its use in ice-sheet modeling highlighted the capacity to capture nuanced physical dynamics in glaciological systems.
\\ \\
POU-PINNs also address challenges in traditional PDE modeling by improving accuracy and computational efficiency. The framework reduces solution space complexity, accelerates convergence, and ensures precise domain-specific physics representation. POU-PINNs demonstrate significant potential in solving real-world problems through their applications, offering a versatile, unsupervised learning approach for physics-informed domain decomposition and expert systems. This model opens pathways for further exploration in diverse fields requiring localized solutions to complex physical phenomena.

\FloatBarrier

\section*{Funding}

The National Science Foundation Graduate Research Fellowship Program (NSF GRFP). The Air Force Office of Scientific Research funded this research under the Agile Science of Test and Evaluation (T\&E) program. The U.S. Department of Energy Minority Serving Institutions Partnership Program (DOE-MSIPP) funded this research under grant number GRANT13584020.

\section*{Acknowledgments}
I acknowledge Prof. Nathaniel A. Trask from the University of Pennsylvania, Drs. Mauro Perego, Jonas A. Actor, and Anthony Gruber from Sandia National Laboratories for their guidance in this work.

\end{document}